%% file: main.tex
\documentclass[letterpaper]{article} 
\usepackage{aaai24}  
\usepackage{times}  
\usepackage{helvet}  
\usepackage{courier}  
\usepackage[hyphens]{url}  
\usepackage{graphicx} 
\usepackage{natbib}  
\usepackage{caption} 
\usepackage{booktabs}
\usepackage{amsfonts,amssymb,bm}
\usepackage{amsmath} 
\usepackage{amsthm}  
\usepackage{multirow}
\usepackage{enumitem}
\usepackage{algorithm}
\usepackage{algorithmic}

\usepackage{newfloat}
\usepackage{listings}
\DeclareCaptionStyle{ruled}{labelfont=normalfont,labelsep=colon,strut=off} 
\floatstyle{ruled}
\newfloat{listing}{tb}{lst}{}
\floatname{listing}{Listing}
\title{Dynamic Spiking Framework for Graph Neural Networks}
\def\method{Dy-SIGN}
\author{
    Nan Yin\textsuperscript{\rm 1}, Mengzhu Wang\textsuperscript{\rm 2}, Zhenghan Chen\textsuperscript{\rm 3}, Giulia De Masi\textsuperscript{\rm 4}, Huan Xiong\textsuperscript{\rm 5,1}\thanks{Corresponding authors.}, Bin Gu\textsuperscript{\rm 6,1*} 
}
\affiliations{
    \textsuperscript{\rm 1}Mohamed bin Zayed University of Artificial Intelligence\\
    \textsuperscript{\rm 2}School of Artificial Intelligence, Hebei University of Technology\\
    \textsuperscript{\rm 3}Microsoft Corporation\\
    \textsuperscript{\rm 4}Technology Innovation Institute\\ \textsuperscript{\rm 5}Harbin Institute of Technology\\
    \textsuperscript{\rm 6}Jilin University\\ 
    \{yinnan8911,dreamkily,pandaarych,jsgubin,\\huan.xiong.math\}@gmail.com, giulia.demasi@tii.ae

}

\begin{document}
\maketitle
\input{0_abstract}
\input{1_introduction}
\input{2_related_work}
\input{3_preliminary}

\input{4_method}

\input{5_experiment}

\input{6_conclusion}

\section{Acknowledgments}
This work is part of the research project ``ENERGY-BASED
PROBING FOR SPIKING NEURAL NETWORKS", performed at Mohamed bin Zayed University of Artificial Intelligence (MBZUAI), funded by Technology Innovation Institute (TII) (Contract No. TII/ARRC/2073/2021).

\bibliography{aaai24}

\newpage
\input{7_appendix}

\end{document}

%% file: 0_abstract.tex
\begin{abstract}

The integration of Spiking Neural Networks (SNNs) and Graph Neural Networks (GNNs) is gradually attracting attention due to the low power consumption and high efficiency in processing the non-Euclidean data represented by graphs. 
However, as a common problem, dynamic graph representation learning faces challenges such as high complexity and large memory overheads. Current work often uses SNNs instead of Recurrent Neural Networks (RNNs) by using binary features instead of continuous ones for efficient training, which would overlooks graph structure information and leads to the loss of details during propagation.  
Additionally, optimizing dynamic spiking models typically requires propagation of information across time steps, which increases memory requirements. To address these challenges, we present a framework named \underline{Dy}namic \underline{S}p\underline{i}king \underline{G}raph \underline{N}eural Networks (\method{}). To mitigate the information loss problem, \method{} propagates early-layer information directly to the last layer for information compensation. To accommodate the memory requirements, we apply the implicit differentiation on the equilibrium state, which does not rely on the exact reverse of the forward computation. While traditional implicit differentiation methods are usually used for static situations, \method{} extends it to the dynamic graph setting. Extensive experiments on three large-scale real-world dynamic graph datasets validate the effectiveness of \method{} on dynamic node classification tasks with lower computational costs.

\end{abstract}

%% file: 1_introduction.tex
\section{Introduction}

Graph Neural Networks (GNNs)~\cite{scarselli2008graph} have been widely applied in various fields to learn the graph representation by capturing the dependencies of nodes and edges, such as relation detection~\cite{schlichtkrull2018modeling,9201410,mi2020hierarchical,xu2019spatial,ju2024survey} and recommendation~\cite{wu2022graph}.
However, most GNNs are primarily designed for static or non-temporal graphs, which cannot meet the requirement of dynamic evolution over time in practice. The established solution is to extend the dynamic graphs into sequence models directly. Typically, these methods~\cite{shi2021gaen,kumar2019predicting} utilize the Recurrent Neural Networks (RNNs)~\cite{cho2014learning} to capture the dynamic evolution of graphs for numerous downstream tasks such as time series prediction or graph property prediction~\cite{wieder2020compact}.

Despite the promising performance of dynamic graphs, the majority of these approaches typically involve complex structures that consume significant computational resources during training and testing. 
Inspired by the way the brain process information, Spiking Neural Networks (SNNs) represent the event or clock-driven signals as inference for updating the neuron nodes parameters~\cite{brette2007simulation}. Different from traditional deep learning methods, SNNs utilize discrete spikes information instead of continuous features, resulting in significantly lower power consumption during model training. Considering the inherent characteristics of SNNs~\cite{maass1997networks,pfeiffer2018deep,schliebs2013evolving}, a few recent works~\cite{zhu2022spiking,xu2021exploiting,li2023scaling} have attempted to integrate SNNs into the GNNs framework to tackle the issue of high computational complexity. These methods transform the node features into a series of spikes with Poisson rate coding, and follow a graph convolution layer with SNN neurons, which employ a membrane potential threshold to convert continuous features to spike information~\cite{kim2020spiking,bu2022optimized}. Although Spiking Graph Networks (SGNs) are gradually gaining attention, the use of SNNs in dynamic graphs is still less explored, which is a more common scenario in life. To address the gap, the work focuses on the problem of \textit{spiking dynamic graph}, which applies SNNs for dynamic graph node classification. 

However, the problem is highly challenging due to the following reasons: (1) Information Loss. The representation of GNNs includes the information on graph structure and neighboring nodes, which are crucial for downstream tasks. However, SNNs employ spike signals instead of continuous features, leading to the loss of details regarding the structure and neighbors. Moreover, with the evolution of graphs over time, the information loss issue may further deteriorate the graph representation. (2) Memory Consumption. The RNN-based dynamic graph methods typically require significant memory resources to store the temporal node information~\cite{sak2014long}. Moreover, SNNs inherently operate with multiple time latencies (i.e., calculate the spike signals with time latency steps in each SNN layer). If we simply replace the GNN layer with the SNN layer on each time step, we need to store the temporary spikes in SNN layer and temporal information at each time step simultaneously, which further exacerbates memory consumption.

In this paper, we present a novel framework named \underline{Dy}namic \underline{S}p\underline{i}king \underline{G}raph \underline{N}eural Networks (\method{}) for node classification. The primary insight of proposed \method{} is to thoroughly explore how to apply SNNs to dynamic graphs, and address the challenges of information loss and memory consumption by using the information compensation mechanism and implicit differentiation on the equilibrium state. On the one hand, the information compensation mechanism aims to make up for the information loss during forward propagation. However, implementing the mechanism in each layer of SNNs would significantly increase the model complexity. Thus, we propose to establish an information channel between the shallow and final layers to incorporate the original information directly into feature representations. This approach not only reduces the model complexity but also mitigates the impact of information loss. On the other hand, inspired by recent advances in implicit methods~\cite{bai2020multiscale,xiao2021training} that view neural networks as solving an equilibrium equation for fixed points and provide alternative implicit models defined by the equation, we provide a variation training method that is suitable for dynamic spiking graph neural networks. Specifically, \method{} simplifies the non-differentiable items in backpropagation, thus avoiding the huge computational overhead of traditional SNNs due to the use of surrogate learning techniques.
In this way, the calculation of the gradient would significantly reduce memory consumption.
We conduct extensive experiments to demonstrate the effectiveness of proposed \method{} in comparison to the state-of-the-art methods across various scenarios.

In summary, our contributions are as follows: 
(1) \textit{Motivation:} From the perspective of practical application and data analysis, we propose the \method{}, which is the first attempt to introduce implicit differentiation into dynamic graph. (2) \textit{Methodology:} We propose a novel approach called \method{} that incorporates SNNs into dynamic graphs to release the information loss and memory consumption problem. (3) \textit{Experiments:} Extensive experiments validate the superiority of the proposed \method{} over the state-of-the-art methods.



%% file: 2_related_work.tex
\section{Related Work}

\subsection{SNN Training Methods}

In the supervised training of SNNs, there are two primary research directions. 
One direction focuses on establishing a connection between the spike representations of SNNs, such as firing rates, and equivalent Artificial Neural Networks (ANNs) activation mappings. This connection enables the ANN-to-SNN conversion~\citep{diehl2015fast,hunsberger2015spiking,rueckauer2017conversion,rathi2020enabling}, and the optimization of SNNs using gradients computed from this equivalent mappings~\cite{thiele2020spikegrad,wu2021training,zhou2021temporal,xiao2021training,meng2022training-ICCV}.  These methods usually require a relatively large number of time-steps to achieve performance comparable to ANNs, suffering from high latency and usually more energy consumption. 
The other direction is to directly train SNNs with back-propagation
~\citep{bohte2000spikeprop,esser2015backpropagation,bellec2018long,huh2018gradient}, which typically employs the surrogate gradients~\citep{shrestha2018slayer} method to overcome the non-differentiable nature of the binary spiking functions and direct train SNNs from scratch. This follows the backpropagation through time (BPTT) framework. BPTT with surrogate gradients can achieve extremely low latency, however, it requires large training memory to maintain the computational graph unfolded over time.

\vspace{-0.05in}
\subsection{Dynamic GNNs}
Dynamic GNNs have achieved impressive performance in various tasks. With the help of RNNs~\cite{cho2014learning}, static GNNs can be extended to model dynamic processes by employing RNN architectures~\cite{rossi2020temporal,pareja2020evolvegcn,shi2021gaen,rossi2020temporal,xu2019spatio}. TGN~\cite{rossi2020temporal} and JODIE~\cite{kumar2019predicting} update the node hidden state by RNN units for representation learning. EvolveGCN~\cite{pareja2020evolvegcn} uses the RNN to regulate the model parameters on each time step. However, the RNN-based dynamic graph methods could save the historical information for graph representation, they typically require massive computational costs and memory consumption. To effectively model the dynamic evolution of graphs while minimizing computational and memory requirements, we introduce the implicit models and SNNs into dynamic GNNs.

\vspace{-0.05in}
\subsection{Feedback Models}

Implicit models are promising approaches to deep learning that utilize implicit layers to determine the outputs.  
In contrast to explicit models, which typically require storing intermediate activations for backpropagation, implicit models use the fixed-point solution~\cite{bai2019deep,bai2020multiscale} to perform backpropagation without saving these intermediate activations. This results in constant complexity for the implicit models, which is a significant advantage for large models.
DEQ~\cite{bai2019deep} demonstrates the ability of implicit models in sequence modeling. 
MDEQ~\cite{bai2020multiscale} incorporates multiscale modeling into implicit deep networks, enabling tasks such as image classification and semantic segmentation. 
To further enhance the efficiency of implicit models, 
~\cite{gu2020implicit} extend the concept of implicit fixed-point equilibrium to graph learning, to address the problem of evaluation and training for recurrent GNNs. ~\cite{liu2022mgnni} propose a multiscale graph neural network with implicit layers to model multiscale information on graphs at multiple resolutions. 
Although implicit models have shown promise in various areas, their application to dynamic spiking graphs is still relatively unexplored. 

%% file: 3_preliminary.tex
\section{Preliminary}

\subsection{Spiking Neuron Models}

SNNs utilize binary activations in each layer, which limits the representation capacity. To address the issue, SNNs introduce a temporal dimension, known as latency $K$. In the forward pass of SNNs, inputs are presented as streams of events and repeated for $K$ time steps to produce the final result. The leaky-integrate-and-fire (LIF) model is commonly used to describe the dynamics of spiking neurons. In LIF, each neuron integrates the received spikes as the membrane potential $\bm{u}_{\tau,i}$, which can be formulated as a first-order differential equation, 
\begin{equation}
    \textbf{LIF: } \bar{\lambda} \frac{d u_\tau}{d\tau} = -(u_\tau - u_{rest}) R \cdot I(\tau), \quad u_\tau < V_{th}\ ,
\end{equation}
where $I(\tau)$ is the input current, $V_{th}$ is the spiking threshold, and $R$ and $\bar{\lambda}$ are resistance and time constant, respectively. When $u_\tau$ reaches $V_{th}$ at time $\tau$, a spike is generated and $u_\tau$ is reset to the resting potential $u_\tau=u_{rest}$, which is usually taken as 0. The spike train is expressed by the Dirac delta function: $s_\tau = \sum_{t^f} \delta(\tau-t^f)$. We consider a simple current model 
$ I_{\tau,i} = \sum_{j} w_{ij} s_{\tau,j} + b ,  $
where $w_{ij}$ is the weight from neuron $j$ to neuron $i$. Then, the general form of LIF is described as:
\begin{equation}
\label{LIF}
\left\{
    \begin{array}{lr}
    u_{\tau+1,i} = \lambda (u_{\tau,i} - V_{th} s_{\tau,i}) + \sum_{j} w_{ij} s_{\tau,j} + b , \\
    s_{\tau+1,i} = \mathbb{H}(u_{\tau+1,i} - V_{th}), 
    \end{array}
\right.
\end{equation}
where $\mathbb{H}(x)$ is the Heaviside step function, which is the non-differentiable spiking function. $s_{\tau,i}$ is the binary spike train of neuron $i$, and $\lambda<1$ is a leaky term related to the constant $\tau_m$ and discretization time interval used in the LIF model. The constant $R$, $\bar{\lambda}$, and time step-size are absorbed into the weights $w_{ij}$ and bias $b$. 
The training of SNNs follows the process of BPTT, and the gradients with $K$ time latency steps are calculated with:
\begin{equation}
\label{gradient}
\begin{aligned}
    \frac{\partial \mathcal{L}}{\partial \bm{W}^l} &= \sum_{\tau=1}^K \frac{\partial \mathcal{L}}{\partial \bm{s}_\tau^{l+1}} \frac{\partial \bm{s}_\tau^{l+1}}{\partial \bm{u}_\tau^{l+1}}\Bigg(\frac{\partial \bm{u}_\tau^{l+1}}{\partial \bm{W}^l} \\&+ \sum_{k<\tau} \prod_{i=\tau-1}^k\left(\frac{\partial \bm{u}_{i+1}^{l+1}}{\partial \bm{u}_{i}^{l+1}}+\frac{\partial \bm{u}_{i+1}^{l+1}}{\partial \bm{s}_i^{l+1}}\frac{\partial \bm{s}_i^{l+1}}{\partial \bm{u}_i^{l+1}}\right)\frac{\partial \bm{u}_k^{l+1}}{\partial \bm{W}^l}\Bigg),
\end{aligned}
\end{equation}
where $\bm{W}^l$ is the trainable matrix on $l$-th layer and $\mathcal{L}$ is the loss. The terms $\frac{\partial \bm{s}_\tau^l}{\partial \bm{u}_\tau^l}$ are non-differentiable, and surrogate derivatives are typically used instead.

\subsection{Dynamic GNNs}

Given the dynamic graph $\mathcal{G}=\{G_1,\cdots,G_t,\cdots,G_T\}$ with $T$ time steps. On each snapshot $G_t=(\bm{A}_t,\bm{X}_t,\mathcal{V}_t,\mathcal{E}_t)$, where $\bm{A}_t$ is the adjacency matrix, $\bm{X}_t\in \mathbb{R}^{N\times d}$ is $N$ node features with dimension $d$, $\mathcal{V}_t=\{v_1^t,\cdots,v_N^t\}$ and $\mathcal{E}_t$ are the set of nodes and edges on time step $t$. Dynamic graph methods typically extract the graph features on each time step and then model the evolution over time, which is formulated as:
\begin{equation}
\label{gnn}
\begin{aligned}
    \bm{h}_v^{t,l}=\mathcal{C}^{t,l}&\left(\bm{h}_v^{t,l-1},\mathcal{A}^{t,l}\left(\{\bm{h}_u^{t,l-1}\}_{u\in \mathcal{N}(v)}\right)\right), \\ \bm{h}_v^{t+1}&=Evo(\bm{h}_v^{1,L},\cdots,\bm{h}_v^{t,L}),
\end{aligned}
\end{equation}
where $\mathcal{N}(v)$ is the neighbors of $v$. $\mathcal{A}^{t,l}$ and $\mathcal{C}^{t,l}$ denote the aggregation and combination operations at the $l$-th layer on time step $t$, respectively. $L$ is the number of layers of a graph and $Evo$ means the evolution operation over time $1$ to $t$, which is typically implemented with RNN~\cite{cho2014learning} or LSTM~\cite{hochreiter1997long}.

%% file: 4_method.tex
\section{The Proposed \method{}}

\begin{figure*}[t]
  \centering
  \includegraphics[scale=0.7]{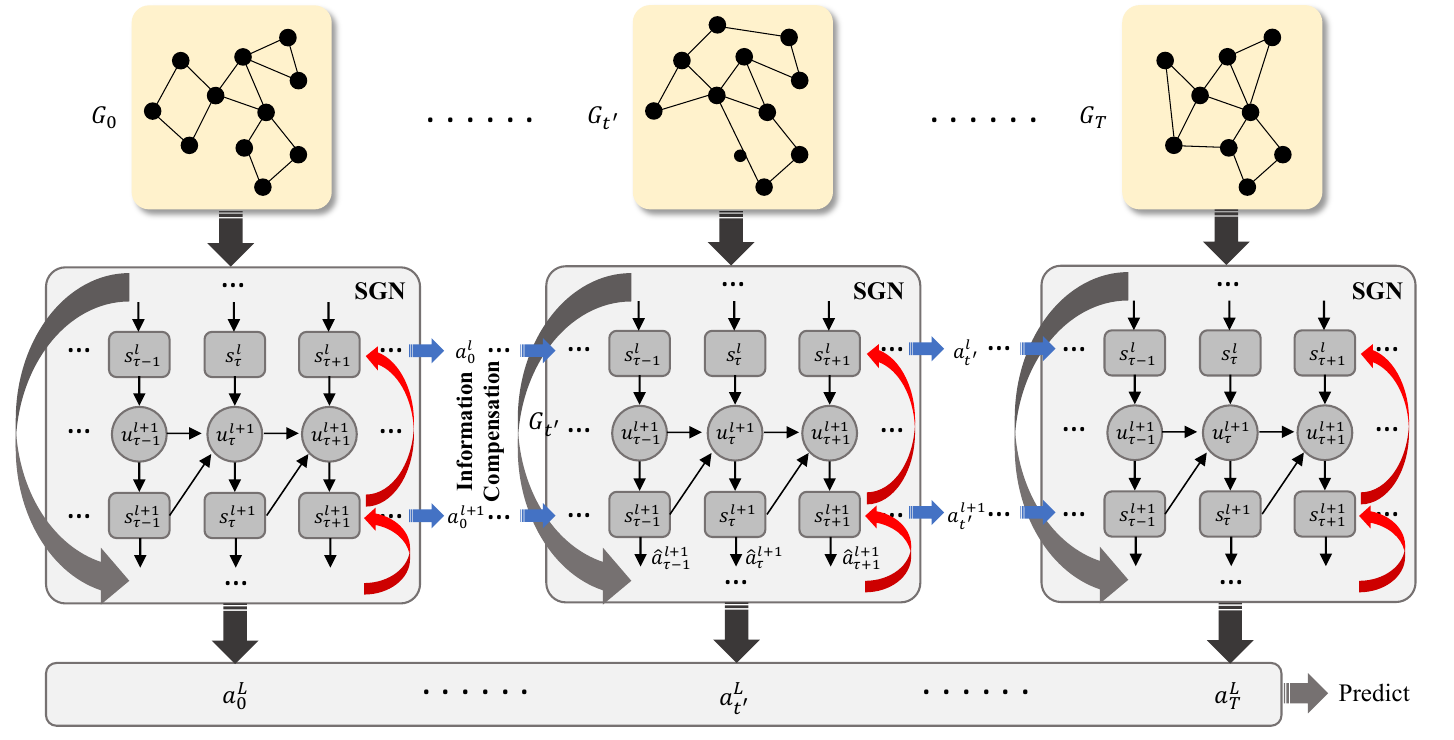}
  \caption{An overview of the proposed \method{}. The Spiking Graph Neural Network (SGN) combines the SNNs and GNNs for node representation learning. The information compensation mechanism transfers the information from the shallow layer to the last to mitigate the information loss issue. The variation training method is applied to calculate the fix-point on each time latency in SGN and is used for dynamic prediction.}
  \label{fig1}
\end{figure*}

\subsection{Overview}
This paper introduces a novel approach named \method{} for semi-supervised dynamic spiking graph node classification. Recognizing that if the time latency $\tau$ in SNN tends to infinity, models will retain the details information of input. However, since the large $\tau$ would cause the vanishing gradient problem~\cite{hochreiter1998vanishing} and significantly increase the complexity, we propose the information compensation mechanism that bridges the feature from the beginning to the last layer and includes the shallow representation in the final embedding. Additionally, we propose a variation of the training method in dynamic graph, which is proved to be equivalent to the implicit differentiation. This method simplifies the calculation of gradient, which relies on the equation of implicit differentiation rather than the forward procedure, thereby reducing memory consumption. The detailed illustration of our \method{} can be seen in Figure \ref{fig1}, and we will introduce \method{} in detail.

\subsection{Information Compensation Spiking Graph Neural Network}
\label{SGN}

Spiking Graph Network (SGN)~\cite{zhu2022spiking,xu2021exploiting} usually applies the Bernoulli encoding to transform the node representation to the spike signals for propagation. Specifically, on time step $t\in[1,\cdots,T]$ with $T$ denotes the length of dynamic graph time window, we have the graph $G_t=(\bm{A}_t,\bm{X}_t)$, where $\bm{A}_t$ is the adjacency matrix of $G_t$ and $\bm{X}_t$ is the node features. SGN first encodes the initial features into binary signals $\{\tilde{\bm{X}}_{t,1},\cdots,\tilde{\bm{X}}_{t,\tau},\cdots,\tilde{\bm{X}}_{t,K}\}$, where $\tau\in[1,\cdots,K]$ means the time latency step in SGN, and $K$ is the length of SGN time latency window. Then, the \textbf{layer-wise} (update on different layers) spiking graph propagation is defined as:
\begin{equation}
\bm{s}_{t,\tau}^l=\Phi\left(Cov(\bm{A},\bm{s}_{t,\tau}^{l-1}),\bm{s}_{t,\tau-1}^l\right),
\end{equation}
where $Cov$ is the graph convolutional layer which is the same as Equation~\ref{gnn}, $S_{t,\tau}^l$ denotes the nodes spiking features on time step $t$ and time latency step $\tau$ with $l$-th layer. $\bm{S}_{t,1}^1=\tilde{\bm{X}}_{t,1}$, $\Phi(\cdot)$ is the non-linear function that combine historical information $\bm{S}_{t,\tau-1}^l$ into current state. After that, the \textbf{temporal-wise} (update on each time latency in SGN) membrane potentials and firing rate follows:
\begin{equation}
\label{snn_update}
\left\{
    \begin{array}{lr}
    \bm{u}_{t,\tau+1}^l=\lambda \bm{u}_{t,\tau}^l(\bm{1}-\bm{s}_{t,\tau}^l)+\sum \bm{W}^l Con(\bm{A},\bm{s}_{t,\tau+1}^{l-1}),\\
    \bm{s}_{t,\tau+1}^l=\mathbb{H}(\bm{u}_{t,\tau+1}^l-V_{th}).\nonumber
    \end{array}
\right.
\end{equation}

For the specific node representation $\bm{x}_{i}=[\bm{x}_{i1},\cdots,\bm{x}_{id}]$, where $\bm{x}_i$ denotes the $i$-th node features of graph, and $d$ is the feature dimension. The spiking signals are sampled with Bernoulli distribution with $K$ time latency in SGN, which is denoted as $\tilde{\bm{x}}_i=\{\tilde{\bm{x}}_{1,i},\cdots,\tilde{\bm{x}}_{\tau,i},\cdots,\tilde{\bm{x}}_{K,i}\}$ with $\tilde{\bm{x}}_{\tau,i}=[\tilde{\bm{x}}_{\tau,i1},\cdots,\tilde{\bm{x}}_{\tau,id}]$. Then we have $P(\tilde{\bm{x}}_{\tau,ij}=1)=\bm{x}_{ij}$ and $P(\tilde{\bm{x}}_{\tau,ij}=0)=1-\bm{x}_{ij}$. Assume the parameters of each spike neuron are $w_i=[w_{i1},\cdots,w_{id}]$, the combined spike input $z_i$ for the next SGN layer holds:
\begin{equation}
\begin{aligned}
    z_{\tau,i}=\sum_{j=1}^d & w_{ij}\tilde{\bm{x}}_{\tau,ij}^\tau, \\ \mathbb{E}(z_{\tau,i})=\sum_{j=1}^d w_{ij}& \mathbb{E}(\tilde{\bm{x}}_{\tau,ij})=\sum_{j=1}^d w_{ij}\bm{x}_{ij}.
\end{aligned}
\end{equation}
According to~\cite{chung2002connected}, the error bound of SGN holds:
\begin{equation}
\label{bound}
\begin{aligned}
     &\lim_{\tau \to\infty} P\left(z_{\tau,i} < \mathbb{E}(z_{\tau,i}) - \epsilon\right) \leq e^{-\epsilon^2/2\sigma},\\
     &\lim_{\tau \to\infty} P\left(z_{\tau,i} > \mathbb{E}(z_{\tau,i}) + \epsilon\right) \leq e^{-\epsilon^2/2(\sigma+\hat{w}_i\epsilon/3)},
\end{aligned}
\end{equation}
where $\hat{w}_i=max\{w_{i1},\cdots,w_{id}\}$. From Equation~\ref{bound}, we observe that as $\tau\to\infty$, the difference between SGN and GNN will be with the probability of $p=e^{-\epsilon^2/2(\sigma+\hat{w}_i\epsilon/3)}$ to exceed the upper and lower bounds. This reveals that the spiking signals would preserve the details information of continuous features when $\tau \to \infty$. However, with the increase of $\tau$, SGN becomes difficult to train and may suffer from the vanishing gradient problem due to the coefficient of $\lambda$ in Equation~\ref{snn_update}. To address the issue, we design the information compensation mechanism for SGN that directly transfers features from the first layer to the last layer for node embeddings. Formally:
\begin{equation}
\label{feedback}
\left\{
    \begin{array}{lr}
    \bm{u}^{1}_\tau = \lambda \bm{u}^{1}_\tau-1 + \bm{W}^1 s^{N}_\tau + \bm{F}^1\bm{x}_\tau- V_{th}s^{1}_\tau , \\
    \bm{u}^{l}_\tau  =\lambda \bm{u}^{l}_{\tau-1} +\bm{F}^{l}s^{l-1}_\tau-V_{th}s^{l}_\tau, \quad l=2,\cdots,N, \nonumber
    \end{array}
\right.
\end{equation}
where $\bm{u}^{l}_\tau$ denotes the neuronal membrane potential at time $\tau$ on $l$-th layer. $\bm{F}^l$ and $\bm{W}^l$ are the trainable parameters on the $l$-th layer, $\bm{F}^1$ is the information compensation matrix. 
In this way, the information compensation SGN follows the form of multi-layer structures of feedback model~\cite{bai2019deep,bai2020multiscale}. This type of structure has several potential advantages: (1) The forward and backward procedures are decoupled, avoiding the problem of non-differentiable spiking functions. (2) Using implicit differentiation in the equilibrium state, we can compute the gradient without saving the exact forward procedure, thus reducing memory consumption.

\subsection{Variation of Training SGN}
\label{variation}
The traditional training method of SGN follows BPTT, which replaces the non-differentiable term $\frac{\partial \bm{s}^{l}_\tau}{\partial \bm{u}^{l}_\tau}$ with the surrogate derivatives in Equation~\ref{gradient}. However, BPTT relies on multiple backpropagation paths, with would consume a large amount of memory. Similarly to ~\cite{xiao2022online}, we set the gradient of the Heaviside step function to 0, which is formulated as $\frac{\partial \bm{u}^{l}_{\tau+1}}{\partial \bm{s}^{l+1}_\tau}\frac{\partial\bm{s}^{l+1}_\tau}{\partial{u}^{l+1}_\tau}=0$. Then, the gradient of parameters $\bm{W}^l$ is:
\begin{equation}
\label{bptt}
\begin{aligned}
    \frac{\partial L}{\partial\bm{W}^l}&=\sum_{\tau=1}^K \frac{\partial L}{\partial \bm{s}^{l+1}_\tau} \frac{\partial \bm{s}^{l+1}_\tau}{\partial \bm{u}^{l+1}_\tau}\left(\sum_{k\leq \tau} \lambda^{\tau-k}\frac{\partial\bm{u}^{l+1}_k}{\partial \bm{W}^l}\right)\\&=\sum_{\tau=1}^K \frac{\partial L}{\partial \bm{s}^{l+1}_\tau} \frac{\partial \bm{s}^{l+1}_\tau}{\partial \bm{u}^{l+1}_\tau}\left(\sum_{k\leq \tau} \lambda^{\tau-k}\bm{s}^{l}_k\right).
\end{aligned}
\end{equation}
During the forward procedure is $\hat{\bm{a}}^{l}_{\tau+1}=\lambda\hat{\bm{a}}^{l}_\tau+\bm{s}^{l}_{\tau+1}$, where the presynaptic activities can be denoted as $\hat{\bm{a}}^{l}_{\tau+1}=\sum_{k\leq\tau}\lambda^{\tau-k}\bm{s}^{l}_k$. By calculating the gradient of $\frac{\partial L}{\partial \bm{s}^{l+1}_\tau} \frac{\partial \bm{s}^{l+1}_\tau}{\partial \bm{u}^{l+1}_\tau}$, we can directly compute the value of Equation~\ref{bptt} without considering the backpropagation through $\frac{\partial \bm{u}^{l+1}_{\tau+1}}{\partial \bm{u}^{l+1}_\tau}$, which would decrease the complexity and memory consumption of the model. On each time latency $\tau$, the output is denoted as $\hat{\bm{a}}^{l+1}_\tau$, and the output of the SGN is $\{\hat{\bm{a}}^{l}_1,\cdots,\hat{\bm{a}}^{l}_\tau,\cdots,\hat{\bm{a}}^{l}_K\}$ on $l$-th layer.

\subsection{Comparing with Feedback Model}

Note that, in the spiking dynamic graph framework, $t\in [1,\cdots,T]$ stands for the time steps of each graph, and $\tau\in[1,\cdots,K]$ is the time latency in SGN. At each time step $t$, we apply the information compensation SGN to extract the graph features as Equation~\ref{feedback}. In the traditional feedback model, the weighted average firing rate and inputs are denoted as $\bm{a}_{t,K}=\frac{\sum_{\tau=1}^K\lambda^{K-\tau}\bm{s}_{t,\tau}}{\sum_{\tau=1}^K\lambda^{K-\tau}}$ and $\Bar{\bm{x}}_{t,K}=\frac{\sum_{\tau=0}^K\lambda^{K-\tau}\bm{x}_{t,\tau}}{\sum_{\tau=0}^K\lambda^{K-\tau}}$, where $\bm{s}_{t,\tau}$ and $\bm{x}_{t,\tau}$ denote the firing rate and input on time latency $\tau$ in SGN and on time step $t$ of dynamic graph. The LIF model approximate an equilibrium point $\bm{a}^\star_t$ that satisfies $\bm{a}^\star_t=\sigma\left(\frac{1}{V_{th}}(\bm{Wa^\star}_t+\bm{Fx^\star}_t)\right)$. The characteristic is similar to the presynaptic activities $\hat{\bm{a}}^{l}_{t,\tau+1}=\sum_{k\leq\tau}\lambda^{\tau-k}\bm{s}^{l}_{t,k}$. Considering the last time latency $K$ on layer $l$, we have $\hat{\bm{a}}^{l}_{t,K}=\sum_{\tau=1}^K \lambda^{K-\tau}\bm{s}^{l}_{t,\tau}$, which equals to $C\bm{a}_{t,K}^l$ with $C=\sum_{\tau=1}^K\lambda^{K-\tau}$. In other words, the fire rate $\hat{\bm{a}}$ in~\ref{variation} at the last time latency step $K$ on layer $l$ is equivalent to the traditional feedback model.
Thus, we prove that the variation of training SGN is equivalent to the traditional feedback model, and we can calculate the gradient of parameters with Equation~\ref{bptt} directly.

\begin{algorithm}[t]
\caption{Learning Algorithm of \method{}}
\begin{flushleft}
\textbf{Input:} Dynamic graph $\mathcal{G}=\{G_1,\cdots,G_T\}$; Label $\bm{y}$; Network parameters $\bm{\theta}$; Network layers $L$; Time latency of SGN $K$. \\
\textbf{Output}: Trained model parameters $\bm{\theta}$.  \\
\end{flushleft}
\begin{algorithmic}[1] 
\STATE Initialize $\bm{\theta}$.
\STATE \textrm{// \textbf{Forward:}}
\FOR{$t=1,\cdots,T$} 
    \FOR{$l=1,\cdots,L$}
    \STATE Calculate the average firing rate $\bm{a}_{t,K}^l$ with Equation~\ref{dsgn};
    \STATE Collect the fixed point representation $\bm{a}_t^\star$ on layer $L$ and time step $t$;
    \ENDFOR
    \ENDFOR
    \STATE Calculate the output of \method{} $\hat{\bm{y}}$ and the loss $\mathcal{L}$ with Equation~\ref{loss};
    
    \STATE \textrm{// \textbf{Backward:}}
    \FOR{$l=L,\cdots,1$}
    \STATE Calculate the gradient of SGN with Equation~\ref{bptt};
    \STATE Update the parameters $\bm{\theta}$.
\ENDFOR
\end{algorithmic}
\label{algorithm}
\end{algorithm}

\subsection{Dynamic Spiking Graph Neural Network}
The proposed information compensation SGN is designed for a fixed time step $t$. However, since the graphs may change over time, it remains a challenge to integrate the temporal dynamics of SGN with dynamic graphs. We propose a novel method by propagating the medium $\bm{a}_{t,K}=\{\bm{a}_{t,K}^1,\cdots,\bm{a}_{t,K}^l,\cdots,\bm{a}_{t,K}^L\}$ at different time steps. At time step $t+1$, we set the initial membrane potential to $\bm{u}_{t+1,1} =\bm{a}_{t,K}$, and the update process of membrane potentials is:
\begin{equation}
\begin{aligned}
    \bm{u}_{t+1,K} &= \lambda \bm{u}_{t+1,K-1} + \bm{Ws}_{t+1,K-1} \\&+\bm{Fx}_{t+1,K-1}-V_{th}\bm{s}_{t+1,K}.
\end{aligned}
\end{equation}
The average firing rates is defined as $\bm{a}_{t+1,K}=\frac{\sum_{\tau=1}^K \lambda^{K-\tau}\bm{s}_{t+1,\tau}}{\sum_{\tau=1}^K \lambda^{K-\tau}}$, the average inputs as $\bar{\bm{x}}_{t+1,K}=\frac{\sum_{\tau=0}^K \lambda^{K-\tau}\bm{x}_{t+1,\tau}}{\sum_{\tau=0}^K \lambda^{K-\tau}}$, and $\bm{u}_{t+1,1}=\bm{a}_{t,K}$, $\bm{s}_{t+1,1}=\bm{0}$. Then, we have:
\begin{equation}
\label{dsgn}
\begin{aligned}
    \bm{a}_{t+1,K}=&\frac{1}{V_{th}}\Bigg(\frac{\sum_{i=0}^{K-2}\lambda^i}{\sum_{i=0}^{K-1}\lambda^i}\bm{W}\bm{a}_{t+1,K-1}+\bm{F}\bar{\bm{x}}_{t+1,K-1}\\
    &-\frac{\bm{u}_{t+1,K}}{\sum_{i=0}^{K-1}\lambda^i}+\frac{\bm{a}_{t,K}}{\sum_{i=0}^{K-1}\lambda^i}\Bigg)\\
    =&\sigma\Big(\frac{1}{V_{th}}\Big(\frac{\sum_{i=0}^{K-2}\lambda^i}{\sum_{i=0}^{K-1}\lambda^i}\bm{W}\bm{a}_{t+1,K-1}\\&+\bm{F}\bar{\bm{x}}_{t+1,K-1}\Big)\Big)-\frac{\bm{u}_{t+1,K}}{V_{th}\sum_{i=0}^{K-1}\lambda^i}\\&+\frac{\bm{a}_{t,K}}{V_{th}\sum_{i=0}^{K-1}\lambda^i},
\end{aligned}
\end{equation}
where $\sigma(x)=\left\{
  \begin{aligned}
    &1, & x > 1 \\
    &x, & 0 \leq x \leq 1 \\
    &0, &  x < 0
  \end{aligned}
\right.$.
For each time step $t$, we will first calculate the equilibrium point $\bm{a}_{t,K}\to \bm{a}_t^\star$, which is fixed for time step $t+1$. Therefore, the LIF model gradually approximates the equilibrium point $\bm{a}^\star_t$ that satisfies $\bm{a}_t^\star=\sigma\left(\frac{1}{V_{th}}(\bm{W}\bm{a}^\star_t+\bm{Fx}^\star_t)\right)$ with $\bar{\bm{x}}_{t,K}\to \bm{x}^\star_t$.

\begin{table*}[t]
\centering
\setlength{\tabcolsep}{1.3mm}
\begin{tabular}{lccccccccc}
\toprule
\multirow{2}{*}{Methods} & \multicolumn{3}{c}{DBLP} & \multicolumn{3}{c}{Tmall}& \multicolumn{3}{c}{Patent}  \\
\cmidrule(lr){2-4} \cmidrule(lr){5-7} \cmidrule(lr){8-10} 
&40\% &60\% &80\% &40\% &60\% &80\% &40\% &60\% &80\% \\
\midrule
DeepWalk&67.08  &67.17  &67.12 &49.09 &49.29 &49.53  &72.32$\pm$0.9 &72.25$\pm$1.2 &72.05$\pm$1.1\\ 
Node2Vec&66.07 &66.81 &66.93  &54.37 &54.55 &54.58  &69.01$\pm$0.9 &69.08$\pm$0.9 &68.99$\pm$1.0\\ 
HTNE&67.68  &68.24  &68.36  &54.81  &54.89  &54.93   &-  &- &-\\
M$^2$DNE&69.02  &69.48  &69.75  &57.75  &57.99  &58.47  &-  &- &-  \\
DyTriad&60.45  &64.77  &66.42  &44.98  &48.97  &51.16  &-  &- &-  \\
MPNN&64.19$\pm$0.4 &63.91$\pm$0.3  &65.05$\pm$0.5  &47.71$\pm$0.8  &47.78$\pm$0.7  &50.27$\pm$0.5 &-  &- &-  \\ 
JODIE&66.73$\pm$1.0	&67.32$\pm$1.1  &67.53$\pm$1.3  &52.62$\pm$0.8  &54.02$\pm$0.6  &54.17$\pm$0.2   &77.57$\pm$0.8  &77.69$\pm$0.6 &77.67$\pm$0.4  \\ 
EvolveGCN&67.22$\pm$0.3  &69.78$\pm$0.8  &71.20$\pm$0.7  &53.02$\pm$0.7  &54.99$\pm$0.7  &55.78$\pm$0.6  &79.67$\pm$0.4  &79.76$\pm$0.5 &80.13$\pm$0.4 \\
TGAT&\textbf{71.18$\pm$0.4}  &71.74$\pm$0.5  &72.15$\pm$0.3  &56.90$\pm$0.6  &57.61$\pm$0.7  &58.01$\pm$0.7   &81.51$\pm$0.4  &81.56$\pm$0.6  &81.57$\pm$0.5  \\ 
SpikeNet&70.88$\pm$0.4  &71.98$\pm$0.3  &74.64$\pm$0.5  &\textbf{58.84$\pm$0.4}  &\textbf{61.13$\pm$0.8} &\textbf{62.40$\pm$0.6} &83.53$\pm$0.6  &\textbf{83.85$\pm$0.7} &83.90$\pm$0.6   \\ 
\midrule
\method{}  &70.94$\pm$0.1 &\textbf{72.07$\pm$0.1} &\textbf{74.67$\pm$0.5}  &57.48$\pm$0.1 &60.94$\pm$0.2 &61.89$\pm$0.1 &\textbf{83.57$\pm$0.3} &83.77$\pm$0.2 &\textbf{83.91$\pm$0.2}\\
\bottomrule
\end{tabular}
\caption{Macro-F1 score comparisons on three dynamic graph datasets with different training ratios. The results are averaged over five runs, and the best results are in boldface. - denotes time-consuming.}
\label{table_1}
\end{table*}

\begin{table*}[t]
\centering
\setlength{\tabcolsep}{1.3mm}
\begin{tabular}{lccccccccc}
\toprule
\multirow{2}{*}{Methods} & \multicolumn{3}{c}{DBLP} & \multicolumn{3}{c}{Tmall}& \multicolumn{3}{c}{Patent}  \\
\cmidrule(lr){2-4} \cmidrule(lr){5-7} \cmidrule(lr){8-10} 
&40\% &60\% &80\% &40\% &60\% &80\% &40\% &60\% &80\% \\
\midrule
DeepWalk&66.53  &66.89  &66.38 &57.11 &57.34 &57.88  &71.57$\pm$1.3 &71.53$\pm$1.0 &71.38$\pm$1.2\\ 
Node2Vec&66.80 &67.37 &67.31  &60.41 &60.56 &60.66  &69.01$\pm$0.9 &69.08$\pm$0.9 &68.99$\pm$1.0\\ 
HTNE&67.68  &68.24  &68.36  &54.81  &54.89  &54.93   &-  &- &-\\
M$^2$DNE&69.23  &69.47  &69.71  &64.21  &64.38  &64.65  &-  &- &-  \\
DyTriad&65.13  &66.80  &66.95  &53.24  &56.88  &60.72  &-  &- &-  \\
MPNN&65.72$\pm$0.4 &66.79$\pm$0.6  &67.74$\pm$0.3  &57.82$\pm$0.7  &57.66$\pm$0.5  &58.07$\pm$0.6 &-  &-  &-  \\ 
JODIE&68.44$\pm$0.6	&68.51$\pm$0.8  &68.80$\pm$0.9  &58.36$\pm$0.5  &60.28$\pm$0.3  &60.49$\pm$0.3   &77.64$\pm$0.7  &77.89$\pm$0.5  &77.93$\pm$0.4  \\ 
EvolveGCN&69.12$\pm$0.8  &70.43$\pm$0.6  &71.32$\pm$0.5  &59.96$\pm$0.7 &61.19$\pm$0.6  &61.77$\pm$0.6  &79.39$\pm$0.5  &79.75$\pm$0.3 &80.01$\pm$0.3 \\
TGAT&71.10$\pm$0.2  &71.85$\pm$0.4  &73.12$\pm$0.3  &62.05$\pm$0.5  &62.92$\pm$0.4  &63.32$\pm$0.7   &80.79$\pm$0.7  &80.81$\pm$0.6 &80.93$\pm$0.6  \\ 
SpikeNet&\textbf{71.98$\pm$0.3}  &72.35$\pm$0.8  &74.86$\pm$0.5  &\textbf{63.52$\pm$0.7}  &\textbf{64.84$\pm$0.4} &\textbf{66.10$\pm$0.3} &83.48$\pm$0.8  &\textbf{83.80$\pm$0.7} &83.88$\pm$0.9   \\ 
\midrule 
\method{}&71.90$\pm$0.1 &\textbf{72.61$\pm$0.4} &\textbf{74.96$\pm$0.2}  &62.93$\pm$0.3 &64.10$\pm$0.3 &65.82$\pm$0.2 &\textbf{83.50$\pm$0.2} &83.47$\pm$0.1 &\textbf{83.90$\pm$0.2} \\
\bottomrule
\end{tabular}
\caption{Micro-F1 score comparisons on three dynamic graph datasets with different training ratios. The results are averaged over five runs, and the best results are in boldface. - denotes time-consuming.}
\label{table_2}
\end{table*}

Finally, we have fixed point representation over time, i.e., $\bm{a}^\star=\{\bm{a}_1^\star,\cdots,\bm{a}_t^\star,\cdots,\bm{a}_T^\star\}$. We concatenate all the embeddings for the final node classification, which is formulated as:
\begin{equation}
\label{loss}
    \hat{\bm{y}}=FC(||_{t=1}^T\bm{a}^\star),\\ \mathcal{L}=-\sum_{r\in \bm{y}_L} \bm{y}_rln\hat{\bm{y}}_r,
\end{equation}
where $||$ denotes the concatenation operation, $FC$ is the fully connect layer, $\hat{\bm{y}}$ is the nodes prediction, $\bm{y}$ is the ground-truth labels, $\bm{y}_L$ means the set of labeled nodes, and $\mathcal{L}$ is the cross-entropy loss.

In summary, the proposed dynamic spiking graph neural network has several advantages. Firstly, $\bm{a}^\star_t$ is the fixed point of each time step on different layers, which can be considered as the hidden embeddings of time step $t$. Compared to traditional RNN-based dynamic graph methods, directly propagating $\bm{a}^\star_t$ over time would reduce the computational cost of calculating the hidden states and lower the model complexity. Secondly, traditional static feedback models usually set the initial states to $0$, which cannot meet the requirement of dynamic graphs. By sending the previous state to the next time step, the model is able to capture the long temporal dependency for prediction. The detailed algorithm is shown in Algorithm~\ref{algorithm}.

%% file: 5_experiment.tex
\begin{figure*}[t]
  \centering
  \includegraphics[scale=0.57]{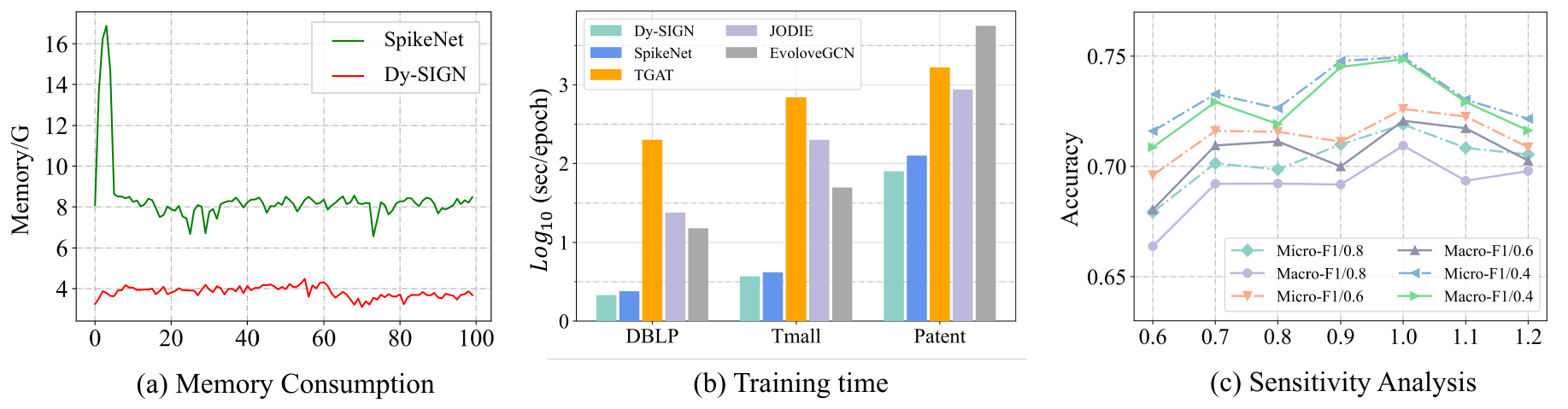}
  \caption{(a) The memory consumption of SpikeNet and \method{} on DBLP dataset. (b) The training time of different methods. (c) Hyperparameter sensitivity analysis of $\lambda$. }
  \label{exp}
\end{figure*}

\section{Experiments}
\subsection{Experimental Settings}
\label{setting}

To verify the effectiveness of the proposed \method{}, we conduct experiments on three large real-world graph datasets, i.e., DBLP~\cite{lu2019temporal}, Tmall~\cite{lu2019temporal} and Patent~\cite{hall2001nber}. The statistics and details introduction are presented in Appendix.
We compared \method{} with various competing methods, including two static graph methods ( i.e., DeepWalk~\cite{perozzi2014deepwalk} and Node2Vec~\cite{grover2016node2vec}), seven dynamic graph methods (i.e., HTNE~\cite{zuo2018embedding}, M$^2$DNE~\cite{lu2019temporal}, DyTriad~\cite{zhou2018dynamic}, MPNN~\cite{panagopoulos2021transfer}, JODIE~\cite{kumar2019predicting}, EvolveGCN~\cite{pareja2020evolvegcn} and TGAT~\cite{Xu2020Inductive}), and one spiking method SpikeNet~\cite{li2023scaling}. The details are introduced in Appendix. As for the implementation,
we follow the same settings with~\cite{li2023scaling} and report the Macro-F1 and Micro-F1 results under different training ratios (i.e., 40\%, 60\%, and 80\%). Besides, we use 5\% for validation. The hidden dimension of all the methods is set to 128, and the batch size to 1024. The total training epochs are 100 and the learning rate is 0.001.

\subsection{Performance Comparison}

\subsubsection{Comparison of Performance.} 
The classification results on different datasets under various training ratios are presented in Table~\ref{table_1} and ~\ref{table_2}. From the results, we find that the proposed \method{} achieves competitive performance compared with other methods. From the results, we have the following observations: (1) The static methods DeepWalk and Node2Vec perform worse than the others, which indicates that simply applying static graph methods to dynamic graphs ignores the contribution of historical information for representation. (2) The methods HTNE, M$^2$DNE, DyTrid, and MPNN fail to learn meaningful representations on the large-scale Patent dataset. The potential reason behind this is the high computational complexity of these models, leading to extensive time consumption for both training and prediction. This makes it impractical to achieve competitive performance within an acceptable time frame.
(3) Although the spiking methods SpikeNet and \method{} apply the binary information for representation learning, the performance is still better than JODIE, EvolveGCN, and TGAT. This phenomenon indicates that the spike-based learning method is also competitive in both computational complexity and performance compared to the traditional methods. (4)  The proposed \method{} outperforms SpikeNet in half of the settings. However, as shown in Figure~\ref{exp} (a), the memory consumption of \method{} is about only half of SpikeNet under the same experiment environment, demonstrating the superiority of \method{}. We attribute this to the fact that by applying the variation training method, \method{} achieves more efficient results.

\subsubsection{Comparison of Runtime Complexity.} We further compare the runtime complexity between \method{} with SpikeNet, JODIE, EvolveGCN, and TGAT, which is shown in Figure~\ref{exp} (b). From the results, we find that the SNN-based methods are significantly more efficient than the ANNs methods. The reason for this is attributed to the fact that SNN-based methods use binary signals instead of continuous features, allowing the matrix multiplication operation to be replaced by an accumulation operation. Additionally, the proposed \method{} is slightly more efficient than SpikeNet method, the potential reason is that \method{} uses the simple form to calculate the gradient with Equation~\ref{bptt}, ignoring the time-consuming calculation of gradient in BPTT.

\subsection{Sensitivity Analysis}
In this section, we examine the impact of hyperparameters on the performance of our proposed \method{}. Specifically, the parameter $\lambda$ determines the amount of information retained for the next time latency step representation. We test the values of $\lambda$ in the range of \{0.6, 0.7, 0.8, 0.9, 1, 1.1, 1.2\} with other parameters fixed to determine the optimal value. The results are depicted in Figure~\ref{exp} (c). From the results, we observe that as the value of $\lambda$ increases, the performance initially improves and then gradually declines. We attribute the reason to the fact that the smaller $\lambda$ cannot provide sufficient historical information for effective representation learning. On the other hand, larger $\lambda$ may lead to worse performance since the current time latency step representation may be influenced by too much historical information. Thus, we set the default value of $\lambda$ to 1.

%% file: 6_conclusion.tex
\section{Conclusion}\label{sec:conclusion}
In this paper, we study the problem of combining SNNs with dynamic graphs using implicit differentiation for node classification and propose a novel method named \method{}. To tackle the issue of information loss on graph structure and details during SNN propagation, we propose an information compensation mechanism. This mechanism passes the original structure and features to the last layer of the network, which then participates in node representation learning. This structure is very similar to traditional feedback models. Based on this, we use explicit differentiation and a variation training method to address the issue of high memory consumption in the combination of SNNs and dynamic graphs. Extensive experiments on real-world large-scale datasets validate the superiority of the proposed \method{}.


%% file: 7_appendix.tex
\appendix



\begin{table}[h]
\caption{The specific statistics of the experimental datasets.}
\centering
\tabcolsep=2.5pt
\begin{tabular}{ccccccc}
\toprule
Dataset  &\#Nodes  &\#Edges &\#Classes &\#Time steps \\
\midrule
DBLP  &28,085  &236,894 &10 &27  \\
Tmall  &577,314 &4,807,545  &5  &186  \\
Patent  &2,738,012  &13,960,811 &6 &25   \\
\bottomrule
\end{tabular}
\label{dataset}
\end{table}

\section{Datasets}
\label{dataset_detail}
In our work, we conduct our experiments on three public large-scale graph datasets, i.e., DBLP~\cite{lu2019temporal}, Tmall~\cite{lu2019temporal} and Patent~\cite{hall2001nber}. The statistics of the datasets are presented in Table~\ref{dataset}, and the details introduction are as follows:

\begin{itemize}[leftmargin=*]
\item \textbf{DBLP.} DBLP~\cite{lu2019temporal} is an academic co-author graph that is derived from the extensive bibliography website. This dataset is widely used in research to study various aspects of academic collaborations and knowledge dissemination within the scientific community. In the DBLP graph, each node represents an author, while edges between nodes indicate collaborative relationships between pairs of authors who have jointly published papers together.
\item \textbf{Tmall.} Tmall~\cite{lu2019temporal} is a bipartite graph that is constructed from sales data gathered in 2014 from the popular e-commerce platform Tmall.com. This dataset captures the interactions between users and items, where each node represents either a user or an item, and each edge corresponds to a purchase transaction along with a timestamp indicating the time of the purchase.
In the Tmall dataset, users are connected to items based on their purchasing behavior. This rich interaction data provides valuable insights into user preferences, item popularity, and the dynamics of e-commerce transactions. The dataset is particularly interesting for studying recommendation systems, personalized marketing, and understanding user-item interactions in an online retail environment.
\item \textbf{Patent.} The Patent dataset~\cite{hall2001nber} is a citation network consisting of United States patents spanning the years 1963 to 1999. This dataset provides a valuable resource for studying the interconnections and relationships between patents in various domains.
Each node in the Patent dataset represents a patent, and the edges between the nodes indicate citation relationships, where one patent cites another. This citation network captures the flow of knowledge and innovation within the patent system. The dataset contains patents from diverse technological domains and covers a wide range of topics and industries.

\end{itemize}

\section{Baselines}
\label{baseline}
In this section, we compared the proposed \method{} with a wide range of competing approaches, the details introduction is as follows:

\textbf{Static Graph Methods.} We compare the proposed \method{} with two widely used static graph methods, i.e., DeepWalk~\cite{perozzi2014deepwalk} and Node2Vec~\cite{grover2016node2vec}. For these baselines, we set the drop rate in the range of \{0.1,0.3,0.5,0.7\} to select the best performance, the embedding size of all methods is set to 128 as default for the sake of fairness.

\begin{itemize}[leftmargin=*]
\item \textbf{DeepWalk.} DeepWalk~\cite{perozzi2014deepwalk} performs random walks to generate an ordered sequence of nodes from a static graph to create contexts for each node, then applies a skip-gram model to these sequences to learn representations.DeepWalk~\cite{perozzi2014deepwalk} is a graph embedding method that leverages random walks to capture the structural information of a static graph. The main idea behind DeepWalk is to generate sequences of nodes by performing random walks on the graph and then learn node representations from these sequences.

\item \textbf{Node2Vec.} Node2Vec~\cite{grover2016node2vec} is an extension of the DeepWalk algorithm that incorporates biased random walks to explore the neighborhood of nodes and learn representations on a static graph. Node2Vec has been widely used for various graph analysis tasks, such as node classification, link prediction, and visualization. By incorporating biased random walks, Node2Vec can capture both the local and global structural characteristics of a graph, providing more informative and expressive node embeddings.
\end{itemize}

\textbf{Dynamic Graph Methods.} We also compare the proposed \method{} with eight dynamic graph methods, i.e., HTNE~\cite{zuo2018embedding}, M$^2$DNE~\cite{lu2019temporal}, DyTriad~\cite{zhou2018dynamic}, MPNN~\cite{panagopoulos2021transfer}, JODIE~\cite{kumar2019predicting}, EvolveGCN~\cite{pareja2020evolvegcn}, TGAT~\cite{Xu2020Inductive} and SpikeNet~\cite{li2023scaling}. We implement all the models with the codes provided by the corresponding paper. We search the hyperparameters by ranging the learning rate over \{0.01,0.005,0.001, 0.0005, 0.0001\} and dropout rate over \{0.3,0.5,0.7\} and report the best reports. For fairness, we set the embedding size of all the models to 128 as default.
\begin{itemize}[leftmargin=*]
\item \textbf{HTNE.} HTNE~\cite{zuo2018embedding} is a network embedding method that combines the Hawkes process and attention mechanism to capture the formation sequences of neighborhoods in a network. By integrating the Hawkes process and attention mechanism, HTNE can effectively capture the temporal dynamics and interaction patterns in network neighborhoods. This enables the learned embeddings to better represent the evolving nature of network structures and facilitate various downstream tasks such as link prediction, community detection, and anomaly detection.

\item \textbf{M$^2$DNE.} M$^2$DNE~\cite{lu2019temporal} is a network embedding method that specifically focuses on capturing the structural and temporal properties of evolving graphs. It achieves this by incorporating a temporal attention point process and a general dynamics equation into the embedding process.

\item \textbf{DyTriad.} DyTriad~\cite{zhou2018dynamic} is a dynamic network modeling approach that combines the modeling of structural information and evolution patterns based on the triadic closure process. By incorporating both aspects, DyTriad is able to effectively capture the dynamics of evolving graphs.

\item \textbf{MPNN.} MPNN~\cite{panagopoulos2021transfer} is a time-series variant of the message-passing neural network (MPNN) that incorporates a two-layer LSTM (Long Short-Term Memory). MPNN is designed to capture long-range temporal dependencies in temporal graphs by encoding the dynamics into the node representations.

\item \textbf{JODIE.} JODIE~\cite{kumar2019predicting} utilizes a dual Recurrent Neural Network (RNN) architecture to update node embeddings in an evolving graph based on observed interactions. This approach enables the prediction of future embedding trajectories.

\item \textbf{EvolveGCN.} EvolveGCN~\cite{pareja2020evolvegcn} is a method that incorporates Recurrent Neural Networks (RNNs) to dynamically evolve the parameters of Graph Neural Networks (GNNs) over time. By adapting the GNN parameters at each time step, EvolveGCN effectively captures the evolving dynamics of a sequence of graphs.

\item \textbf{TGAT.} TGAT~\cite{Xu2020Inductive} is an inductive learning method specifically designed for dynamic graph data. It leverages self-attention mechanisms as fundamental building blocks and introduces a functional time encoding scheme to effectively capture both temporal and topological information in dynamic graphs. This approach has demonstrated state-of-the-art performance across various dynamic graph learning tasks.

\item \textbf{SpikeNet.} SpikeNet~\cite{li2023scaling} introduces a scalable framework designed to effectively capture both the temporal dynamics and structural patterns of temporal graphs. This framework addresses the challenge of efficiently analyzing large-scale temporal graph data.
\end{itemize}

\section{Our Method}
We implement the proposed \method{} with PyTorch~\cite{paszke2017automatic} and PyTorch Geometric library~\cite{fey2019fast}. For the implementation, we adopt the identical settings as described in~\cite{li2023scaling} and present the evaluation results in terms of Macro-F1 and Micro-F1 scores using various training ratios, namely 40\%, 60\%, and 80\%. Additionally, we allocate 5\% of the data for validation purposes.
In our implementation, we set the hidden dimension of all methods to 128 to ensure consistency. The batch size is configured as 1024 to balance computational efficiency and model performance. To train the models effectively, we conduct a total of 100 training epochs and employ a learning rate of 0.001.
These settings allow for a fair and consistent evaluation across different methods, enabling us to assess their performance under comparable conditions. By reporting Macro-F1 and Micro-F1 scores, we can comprehensively evaluate the models' performance in terms of both overall and individual class classification accuracy.
Overall, our implementation follows a standardized approach, ensuring reproducibility and facilitating meaningful comparisons between different methods.

%